\documentclass[runningheads]{llncs}
\usepackage{graphicx}

\usepackage{tikz}
\usepackage{comment}
\usepackage{amsmath,amssymb} 
\usepackage{color}

\usepackage[accsupp]{axessibility}  

\usepackage{graphicx}
\usepackage{amsmath}
\usepackage{amssymb}
\usepackage{booktabs}
\usepackage{algorithm}  
\usepackage{algorithmicx}
\usepackage{algpseudocode}  
\usepackage{multirow}

\newcommand{\etal}{\emph{et al.}}
\newcommand{\eg}{\emph{e.g. }}
\newcommand{\ie}{\emph{i.e. }}
\usepackage[pagebackref,breaklinks,colorlinks]{hyperref}

\begin{document}
	\pagestyle{headings}
	\mainmatter
	\def\ECCVSubNumber{100}  
	
	\title{LiDAR Distillation: Bridging the Beam-Induced Domain Gap for 3D Object Detection} 

	\titlerunning{LiDAR Distillation}
	%
	\author{Yi Wei$^{1,2}$, Zibu Wei$^1$, Yongming Rao$^{1,2}$, Jiaxin Li$^3$, Jie Zhou$^{1,2}$, Jiwen Lu$^{1,2}$\thanks{Corresponding author}}
	\authorrunning{Yi Wei et al.}
	%
	\institute{$^1$Department of Automation, Tsinghua University, China\\
		$^2$Beijing National Research Center for Information Science and Technology, China\\
		$^3$Gaussian Robotics, China
		\\\email{y-wei19@mails.tsinghua.edu.cn, weizb18@mails.tsinghua.edu.cn,raoyongming95@gmail.com\\ lijx1992@gmail.com, jzhou@tsinghua.edu.cn,lujiwen@tsinghua.edu.cn}}
	
	
	\maketitle
	
	\begin{abstract}
		In this paper, we propose the LiDAR Distillation to bridge the domain gap induced by different LiDAR beams for 3D object detection. In many real-world applications, the LiDAR points used by mass-produced robots and vehicles usually have fewer beams than that in large-scale public datasets. Moreover, as the LiDARs are upgraded to other product models with different beam amount, it becomes challenging to utilize the labeled data captured by previous versions' high-resolution sensors. Despite the recent progress on domain adaptive 3D detection, most methods struggle to eliminate the beam-induced domain gap. We find that it is essential to align the point cloud density of the source domain with that of the target domain during the training process. Inspired by this discovery, we propose a progressive framework to mitigate the beam-induced domain shift. In each iteration, we first generate low-beam pseudo LiDAR by downsampling the high-beam point clouds. Then the teacher-student framework is employed to distill rich information from the data with more beams. Extensive experiments on Waymo, nuScenes and KITTI datasets with three different LiDAR-based detectors demonstrate the effectiveness of our LiDAR Distillation. Notably, our approach does not increase any additional computation cost for inference. Code is available at \href{https://github.com/weiyithu/LiDAR-Distillation}{\color{cyan}{https://github.com/weiyithu/LiDAR-Distillation}}. 
		
	\end{abstract}

	\section{Introduction}
	\label{sec:intro}

	\begin{figure}[tb]
		\centering
		\includegraphics[width=0.7\linewidth]{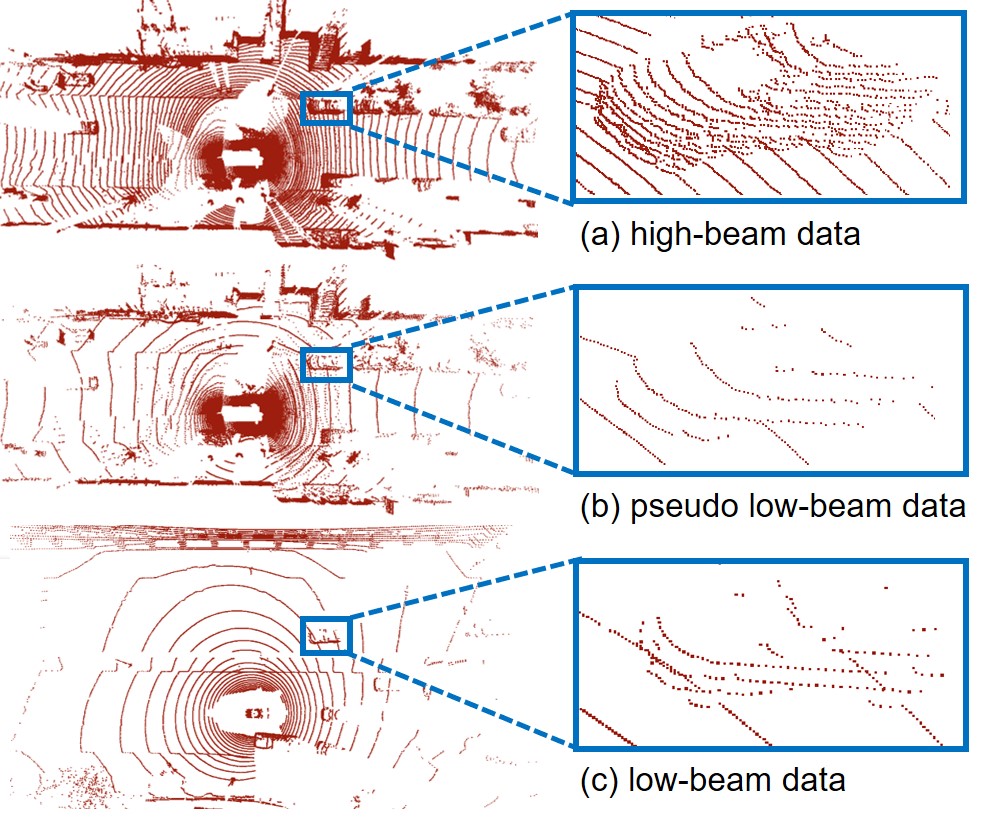}
		\caption{The beam-induced domain gap. The high-beam data (source domain) and the low-beam data (target domain) are from Waymo \cite{sun2020scalability} and nuScenes\cite{caesar2020nuscenes} datasets respectively. We generate pseudo low-beam data by downsampling high-beam data to align the point cloud density of the source domain to the target domain. The models are trained on this synthetic data with a teacher-student framework.}
		\label{fig:pipeline}
	\end{figure}
	
	\begin{figure}[tb]
		\centering
		\includegraphics[width=0.9\linewidth]{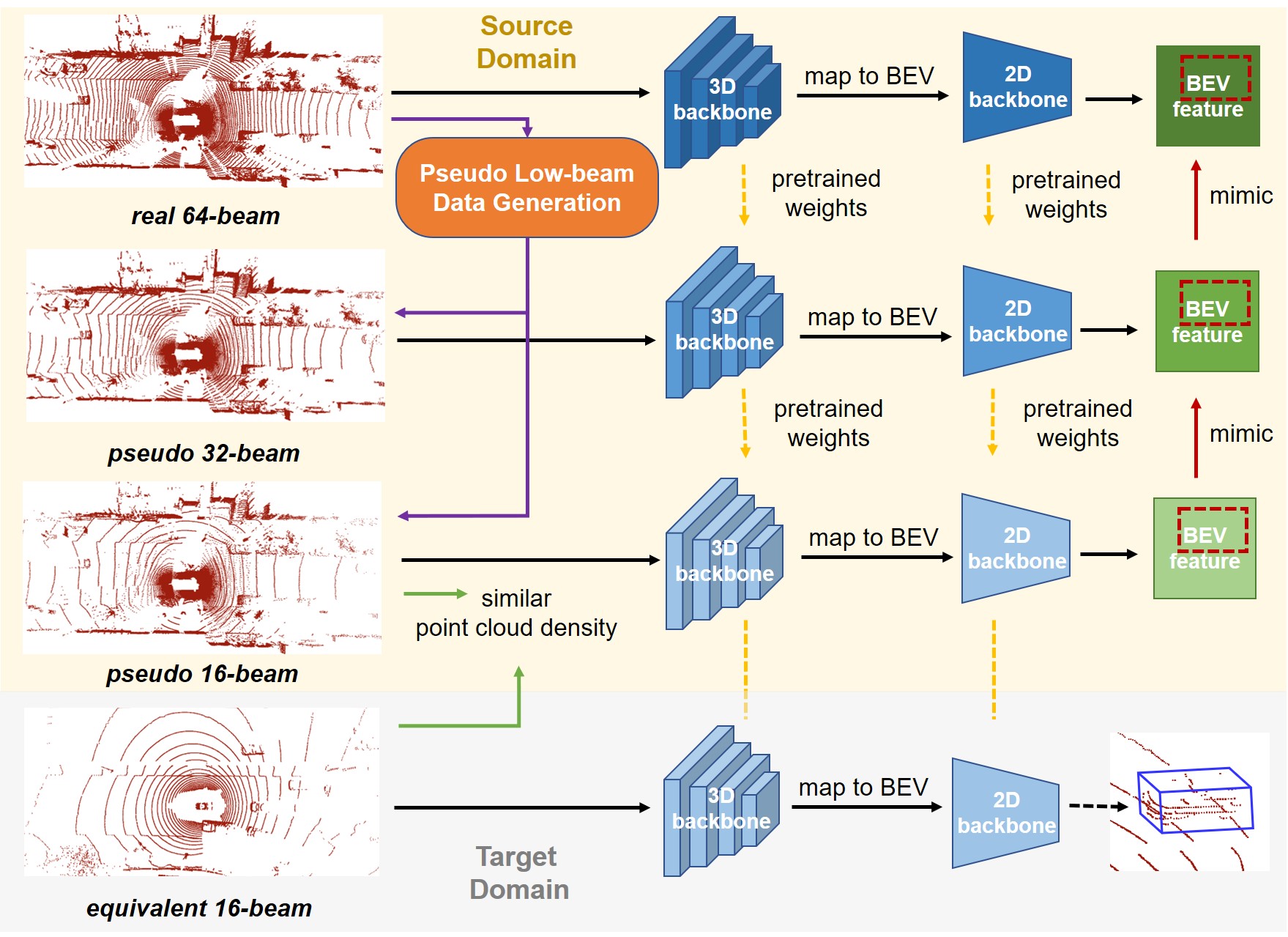}
		\caption{An overview of LiDAR Distillation. Our method aligns the point cloud density of the source domain with that of the target domain in an progressive way. In each iteration, we first generate pseudo low-beam data by downsampling high-beam point clouds. Then we employ a teacher-student framework to improve the model trained on synthetic low-beam data. Specifically, teacher and student networks are trained on high-beam and low-beam data respectively, which have the same architecture. The student network is initialized with the pretrained weights of the teacher model. As summarized in \cite{openpcdet2020}, many 3D detectors employ 3D backbones and 2D backbones, and predict dense BEV features. Based on this fact, we conduct the mimicking operation on the ROI regions of BEV feature maps. }
		\label{fig:overview}
	\end{figure}
	
	Recently, LiDAR-based 3D object detection \cite{lang2019pointpillars,shi2019pointrcnn,shi2019points,shi2020pv,yan2018second,zhou2018voxelnet} has attracted more and more attention due to its crucial role in autonomous driving. While the great improvement has been made, most of the existing works focus on the single domain, where the training and testing data are captured with the same LiDAR sensors. In many real-world scenarios, we can only get low-beam (\eg 16 beams) point clouds since high-resolution LiDAR (\eg 64 beams) is prohibitively expensive. For example, the price of a 64-beam Velodyne LiDAR is even higher than a cleaning robot or an unmanned delivery vehicle. However, since 3D detectors \cite{lang2019pointpillars,yan2018second,shi2020pv} cannot simply transfer the knowledge from the high-beam data to the low-beam one, it is challenging to make use of large-scale datasets \cite{Geiger2012KITTI,sun2020scalability,lyft2019} collected by high-resolution sensors. Moreover, with the update of product models, the
	robots or vehicles will be equipped with the LiDAR sensors
	with different beams. It is unrealistic to collect and annotate
	massive data for each kind of product. One solution is to
	use the highest-beam LiDAR to collect informative training
	data and adapt them to the low-beam LiDARs. Thus, it is an important direction to bridge the domain gap induced by different LiDAR beams.

	Since beam-induced domain gap  directly comes from the sensors, it is specific and important among many domain-variant factors. Different with research-guided datasets, the training data in
	industry tends to have similar environments with that during inference, where the environmental domain gaps such as
	object sizes, weather conditions are
	marginal. The main gap comes from the beam difference of LiDARs
	used during training and inference. Moreover, although LiDAR Distillation is designed for beam-induced domain gap, we can easily combine our method with other 3D unsupervised domain adaptation methods to solve general domain gaps.

	While there are several works \cite{luo2021unsupervised,saltori2020sf,wang2020train,xu2021spg,yang2021st3d,zhang2021srdan} that concentrate on the unsupervised domain adaptation (UDA) task in 3D object detection, most of them aim to address general UDA issues. However, due to the fact that most 3D detectors are sensitive to point cloud densities, the beam-induced domain gap cannot be handled well with these methods. As mentioned in ST3D \cite{yang2021st3d}, it is difficult for their UDA method to adapt detectors from the data with more beams to
	the data with fewer beams. Moreover, since the structure of point clouds is totally different from images, the UDA methods \cite{chen2018domain,zou2018unsupervised,saito2017asymmetric,saito2018maximum,saito2019strongweak,zheng2020crossdomain,chen2020harmonizing,xu2020exploring,li2020spatialattention,zhu2019selective} designed for image tasks are not suitable for 3D object detection.

	To address the issue, we propose the LiDAR Distillation to bridge the beam-induced domain gap in 3D object detection task. The key insight is to align the point cloud density and distill the rich knowledge from the high-beam data in a progressive way. First, we downsample high-beam data to pseudo low-beam point clouds, where downsampling operations are both conducted on beams and points in each beam. To achieve this goal, we split the point clouds to each beam via a clustering algorithm.  Then, we present a teacher-student pipeline to boost the performance of the model trained on low-beam pseudo data. Unlike knowledge distillation \cite{hinton2015distilling,romero2014fitnets,romero2014fitnets,chen2017learning,li2017mimicking} used in model compression, we aim to reduce the performance gap caused by data compression.  Specifically, the teacher and student networks have the same structure, and the only difference comes from the different beams of training data. Initialized by the weights of the teacher model, the student network mimics the region of interest (ROI) of  bird’s eye
	view (BEV) feature maps predicted by the teacher model. 
	
	Experimental results on both cross-dataset \cite{sun2020scalability,caesar2020nuscenes} and single-dataset \cite{Geiger2012KITTI} settings demonstrate the effectiveness of LiDAR Distillation. For cross-dataset adaptation \cite{sun2020scalability,caesar2020nuscenes},  although there exists many other kinds of domain gap (\eg object sizes), our method still outperforms state-of-the-art methods \cite{yang2021st3d,wang2020train}. In addition, our method  is complementary to other general UDA approaches \cite{luo2021unsupervised,saltori2020sf,wang2020train,yang2021st3d,zhang2021srdan} and we can get more promising results combining LiDAR Distillation with ST3D \cite{yang2021st3d}. To exclude other domain-variant factors, we further conduct single-dataset experiments on KITTI \cite{Geiger2012KITTI} benchmark. The proposed method significantly improves the direct transfer baseline with a great margin while the general UDA method ST3D \cite{yang2021st3d} surprisingly degrades the performance.

	\section{Related Work}
	\label{sec:related}
	\noindent \textbf{LiDAR-based 3D Object Detection:}
	LiDAR-based 3D object detection \cite{lang2019pointpillars,shi2019pointrcnn,shi2019points,shi2020pv,yan2018second,zhou2018voxelnet,yang2019std,qi2018frustum} focuses on the problem of localizing and classifying objects in 3D space. It has attracted increasing attention in recent works due to its eager demand in computer vision and robotics. As the pioneer works, \cite{chen2017multi,yang2018pixor,ku2018joint} directly project point clouds to 2D BEV maps and adopt 2D detection methods to extract features and localize objects. Beyond these works, as a commonly used 3D detector in the industry due to its good trade-off of efficiency and accuracy, PointPillars \cite{lang2019pointpillars} leverages an encoder to learn the representation of
	point clouds organized in pillars. Recently, voxel-based methods \cite{yan2018second,zhou2018voxelnet,shi2020pv,deng2021voxel} achieve remarkable performances thanks to the development of 3D sparse convolution \cite{graham20183d,choy20194d}. As a classical backbone, SECOND \cite{yan2018second} investigates an improved sparse convolution method embedded with voxel feature encoding layers. Combining voxel-based networks with point-based set abstraction via a two-step strategy, PV-RCNN \cite{shi2020pv} becomes one of the state-of-the-art methods. In our work, we adopt PointPillars \cite{lang2019pointpillars}, SECOND \cite{yan2018second} and PV-RCNN \cite{shi2020pv} to demonstrate the effectiveness of our LiDAR Distillation.

	\noindent \textbf{Unsupervised Domain Adaptation on Point Clouds:}
	The goal of unsupervised domain adaptation \cite{chen2018domain,zou2018unsupervised,saito2017asymmetric,saito2018maximum,saito2019strongweak,zheng2020crossdomain,chen2020harmonizing,xu2020exploring,li2020spatialattention,zhu2019selective,yihan2021learning} is to transfer the model trained on source domain to unlabeled target domains. Recently, some works begin to concentrate on the UDA problem in 3D tasks. \cite{qin2019pointdan,zhou2018unsupervised} explore object-level feature alignment between global features and local features. Among the UDA methods for 3D semantic segmentation \cite{wu2019squeezesegv2,jaritz2020xmuda,yi2021complete}, \cite{yi2021complete} tries to  solve the domain shift induced by different LiDAR sensors. However, their method adopts 3D convolutions for scene completion, which will bring huge additional computation cost. Moreover, the complete scenes can be easily used in 3D segmentation but they are not suitable for 3D object detection. Compared with great development in 3D object detection, only a few methods \cite{luo2021unsupervised,saltori2020sf,wang2020train,xu2021spg,yang2021st3d,zhang2021srdan} have been proposed for the UDA of 3D detection. Wang \etal \cite{wang2020train} conclude that the key factor of the domain gap is the difference in car size and they propose SN to normalize the object sizes of the source and target domains. Yang \etal \cite{yang2021st3d} propose ST3D which redesigns the self-training pipeline in 2D UDA methods for 3D object detection task. While most of these works target at general UDA methods, we find that they struggle to bridge the beam-induced domain gap.

	\noindent \textbf{Knowledge Distillation:}
	Knowledge distillation (KD) \cite{hinton2015distilling,romero2014fitnets,chen2017learning,li2017mimicking} has become one of the most effective techniques for model compression. KD leverages the teacher-student framework and distills the knowledge from teacher models to improve the performance of student models. As one of the pioneer works, Hinton \etal \cite{hinton2015distilling} transfer the knowledge through soft targets. Compared with the KD in classification task \cite{anil2018large,heo2019comprehensive,tung2019similarity,yim2017gift,you2017learning}, compressing object detectors \cite{chen2017learning,li2017mimicking,wei2018quantization,guo2021distilling,wang2019distilling} is more challenging. Chen \etal \cite{chen2017learning} improve the students by mimicking all components, including intermediate features and soft outputs. Beyond this work, Li \etal \cite{li2017mimicking} only mimic the areas sampled from region proposals. Recently, Guo \etal \cite{guo2021distilling} point out that features from background regions are also important. Different from above methods, we utilize knowledge distillation for data compression (high-beam data to low-beam data) instead of model compression. In addition, few works study the problem of knowledge distillation in 3D object detection. Due to the differences in backbones, it is not trivial to directly apply 2D methods to 3D tasks. In our method, we adopt feature imitation on bird’s eye
	view (BEV) feature maps.
	
	\section{Approach}
	\subsection{Problem Statement}
	The goal of unsupervised domain adaptation is to mitigate the domain shift between the source domain and the target domain and maximize the performance on the target domain. Specifically, in this work, we aim to transfer a 3D object detector trained on high-beam source labeled data $\{(X^s_i, Y^s_i)\}^{N_s}_{i=1}$ to low-beam unlabeled target domain $\{X^t_i\}^{N_t}_{i=1}$, where $s$ and $t$ represent source and target domains respectively. The $N_s$ is the samples number of the source domain while $X^s_i$ and $Y^s_i$ mean the $i$th source point cloud and its corresponding label. The 3D bounding box label $Y^s_i$ is parameterized by its center location of $(c_x, c_y, c_z)$, the size $(d_x, d_y, d_z)$, and the orientation $\theta$.

	\subsection{Overview}
	We propose the LiDAR Distillation to bridge the domain gap induced by LiDAR beam difference between source and target domains. Figure \ref{fig:overview} shows the pipeline of our method. A key factor of our approach is to progressively generate low-beam pseudo LiDAR $\{X^m_i\}^{N_s}_{i=1}$ in the source domain, which has the similar density with the point clouds of the target domain. The density indicates the density of beams and point number per beam. Section \ref{sec:pseudo} introduces this part in details. Then we leverage the teacher-student framework to improve the performance of the model trained on pseudo low-beam data, which is discussed in Section \ref{sec:distillation}. Finally, Section \ref{sec:progressive} shows how we progressively distill knowledge from the informative high-beam real point clouds.  
	
	We find that the general UDA method \cite{yang2021st3d} will surprisingly degrade the performance when dealing with the domain shift only caused by LiDAR beams. The observation indicates that it is hard to directly transfer the knowledge from the high-beam source domain to the low-beam target domain. Although subsampling high-beam source domain data to low-beam ones will lose rich data knowledge, it guarantees the point cloud density of the source domain is similar to that of the target domain,  which is more important compared with data information gain. In this work, we focus on the margin between real high-beam data and generated low-beam data without leveraging the training set in the target domain. Therefore, our approach can be easily integrated with other UDA methods by substituting the source domain with the generated middle domain.

	\begin{figure}[tb]
		\centering
		\setlength\tabcolsep{1.0pt} 
		\renewcommand{\arraystretch}{1.0}
		\begin{tabular}{cc}
			{\includegraphics[width=0.3\linewidth]{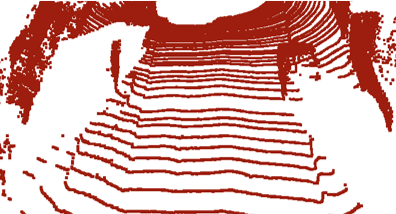}} & {\includegraphics[width=0.3\linewidth]{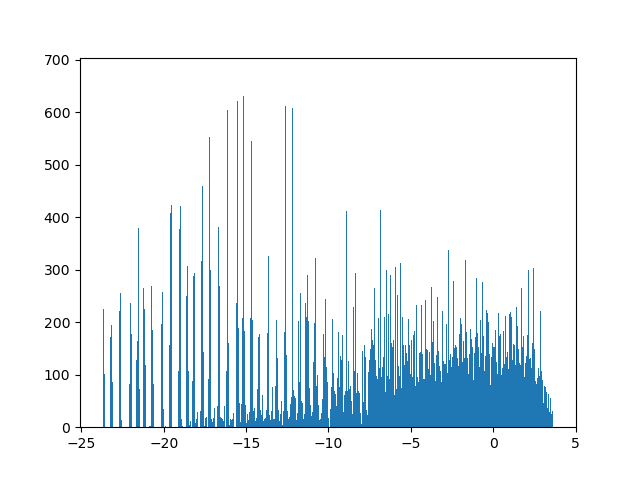}}
			
			\\
			(a) Original data & (b) Zenith angle distribution\\
			{\includegraphics[width=0.3\linewidth]{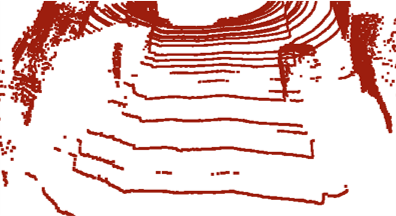}}   & 
			{\includegraphics[width=0.3\linewidth]{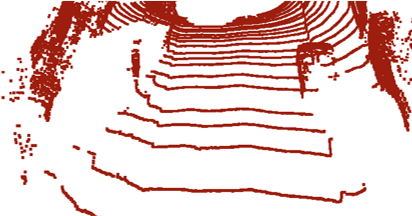}} \\
			(c) Conventional methods & (d) Ours
			
		\end{tabular}
		\centering
		\caption{Pseudo low-beam data generation. \textbf{Top row:} (a) original 64-beam LiDAR points. (b) the distribution of zenith angles $\theta$. \textbf{Bottom row:} (c) pseudo 32-beam data generated by conventional methods \cite{wu2018squeezeseg,milioto2019rangenet++}. (d) pseudo 32-beam data generated by our clustering method. Although we can see the distinct peaks in the distribution of $\theta$, there also exists some noise. Conventional methods will produce short broken lines with the wrong beam labels while our method can generate more realistic data.
		}
		\label{fig:pseudo}
	\end{figure}
	
	\subsection{Pseudo Low-beam Data Generation}
	\label{sec:pseudo}
	The objective of this stage is to close the gap of input point cloud densities between two domains. An alternative solution is to increase the beams of the point clouds in the target domain. This can be easily implemented by upsampling the range images converted from point clouds. However, our experiments show that the upsampled point clouds have many noisy points that heavily destroy objects' features. In this work, we propose to downsample the real high-beam data in the source domain to realize point cloud density's alignment. To this end, we first need to split point clouds into different beams. Although some datasets have beam annotations, many LiDAR points do not have these labels. Here we present a clustering algorithm to get the beam label of each point. We assume the beam number and mean point number in each beam of source and target domains are ${B_s, P_s}$ and ${B_t, P_t}$ respectively. 
	
	We first transfer cartesian coordinates $(x,y,z)$ to spherical coordinates:
	\begin{equation}
		\begin{aligned}
			\theta  = &  \arctan \frac{z}{\sqrt{x^2 + y^2}} \\ 
			\phi  = &  \arcsin \frac{y}{\sqrt{x^2 + y^2}}
		\end{aligned}
	\end{equation}
	where $\phi$ and $\theta$ are azimuth and zenith angles. Figure \ref{fig:pseudo} shows one case in KITTI dataset \cite{Geiger2012KITTI} and the subfigure (b) is the distribution of zenith angles $\theta$. Conventional methods \cite{wu2018squeezeseg,milioto2019rangenet++} assume that beam angles are uniformly distributed in a range (\eg $[-23.6^{\circ}, 3.2^{\circ}]$ in KITTI dataset) and assign a beam label to each point according to the distance between its zenith angle and beam angles.  However, this does not always hold true due to the noise in raw data. To tackle the problem, we apply K-Means algorithm on zenith angles to find $B_s$ cluster centers as beam angles.  As shown in Figure \ref{fig:pseudo} (c)(d), compared to the conventional methods, our method is more robust to the noise and can generate more realistic pseudo data. Moreover, from Table \ref{tab:datasets}, we can see that the vertical field of view (the range of beam angles) of datasets are different.  Intuitively aligning $B_s$ beams to $B_t$ beams is not correct, which will cause different beam densities. We define the equivalent beams $B_t'$ to the source domain as follows:
	\begin{equation}
		\label{eq:beam}
		\begin{aligned}
			B_t' = [\frac{\alpha_s - \beta_s}{\alpha_t - \beta_t} B_t]
		\end{aligned}
	\end{equation}
	where $[\alpha_s, \beta_s]$ and $[\alpha_t, \beta_t]$ represent the vertical field of view in source and target domains. With the beam label of each point, we can easily downsample $B_s$ beams to $B_t'$ beams. Moreover, we should also align $P_s$ to $P_t$ by sampling points in each beam. Instead of random sampling, we sort points according to their azimuth angles $\phi$  and subsample them orderly.

	\subsection{Distillation from High-beam Data}
	\label{sec:distillation}
	Knowledge distillation usually employs a teacher-student framework, where the teacher network is a large model while the student network is a small model. The goal is to improve the performance of the student network by learning the prediction of teacher network. There are two kinds of mimic targets: classification logits and intermediate feature maps. In object detection task, since feature maps are crucial for both object classification and localization, they are the commonly used mimic targets. Formally, regression loss is applied between the feature maps predicted by student and teacher networks:
	\begin{equation}
		L_m = \frac{1}{N} \sum_i ||f_t^i - r(f_s^i)||_2
	\end{equation}
	where $f_t^i$ and $f_s^i$ are the teacher and student networks' feature maps of the $i$th sample. $r$ is a transformation layer to align the dimensions of feature maps. During the mimic training, the weights of teacher are fixed and we only train the student network. 
	
	Different from knowledge distillation methods aiming at model compression, our goal is to distill the knowledge from the high-beam data $\{X^s_i\}^{N_s}_{i=1}$ to the pseudo low-beam data $\{X^m_i\}^{N_s}_{i=1}$, which can be regarded as a data compression problem. Unlike model compression, our student and teacher models employ the same network architecture. The only difference between them is that the input to the student and teacher models are $X^m$ and $X^s$ respectively. To get higher accuracy, we use the pretrained teacher model's parameters as the initial weights of the student network.
	
	
	However, compared with 2D detectors, the architectures of 3D detectors are more diverse. To the best of our knowledge, few works study the mimicking methods for 3D object detection. Moreover, it is not trivial to mimic sparse 3D features since the occupied voxels of student and teacher networks are different. We observe that many 3D object detection methods \cite{lang2019pointpillars,chen2017multi,yang2018pixor,ku2018joint,yan2018second,shi2020pv,deng2021voxel} project 3D features to bird's eye view (BEV). Based on this fact, we conduct mimicking operation on dense BEV features.
	
	As mentioned in \cite{li2017mimicking,wei2018quantization}, the dimension of feature maps can be million magnitude and it is difficult to perform regression of two high-dimension features. This phenomenon also exists in BEV maps. In addition, many regions on feature maps have weak responses and not all elements are crucial for final prediction. The student network should pay more attention to the features in important regions. Here, we propose to mimic the BEV features sampled from the region of interest (ROI). The ROIs contain both positive and negative samples with different ratios and sizes,  which are informative and have significant guidance effects. These ROIs generated by the teacher model will force the student network to learn more representative features. For the 3D detectors which do not use ROIs (\eg PointPillars \cite{lang2019pointpillars} and SECOND \cite{yan2018second}), we can add a light-weight ROI head assigning negative and positive targets during training and remove it during inference. 
	
	The loss function that the student network intends to minimize is defined as follows:
	
	\begin{equation}
		\begin{aligned}
			L &= L_{gt} + \lambda L_m^{3D} \\
			L_m^{3D} = \frac{1}{N} &\sum_i \frac{1}{M_i} \sum_j ||b_t^{ij} - b_s^{ij}||_2
		\end{aligned}
	\end{equation}
	
	where $b^{ij}$ indicates the $j$th ROI of the $i$th sample's BEV feature maps. $N$ and $M_i$ represent sample number and ROI number in the $i$th sample. $L_{gt}$ is the original objective function for 3D object detection and $\lambda$ is the mimic loss weight. Note that there is no need for transformation layer $r$ since student and teacher networks have the same architecture.

	\subsection{Progressive Knowledge Transfer}
	\label{sec:progressive}
	If the beam difference between real high-beam data and pseudo low-beam data is too large (\eg 128 vs 16), the student network cannot learn from the teacher model effectively. To solve the issue, we further present a progressive pipeline to distill the knowledge step by step, which is described in Algorithm \ref{alg:progressive}. For example, if the source domain is 64-beam and the target domain is 16-beam, we first need to generate pseudo 32-beam data and train a student model on it.  This student model will be used as the teacher network in the next iteration. Then we downsample both the beams and points per beam of 32-beam data to 16-beam data. We utilize the student model trained on the generated 16-beam data as the final model for the target domain. We downsample beams for 2 times in each iteration since lots of LiDAR data contains $2^k$  beams.

	\section{Experiments}
	\subsection{Experimental Setup} 
	\noindent \textbf{Datasets:} We conduct experiments on three popular autonomous driving datasets: KITTI \cite{Geiger2012KITTI}, Waymo \cite{sun2020scalability} and nuScenes \cite{caesar2020nuscenes}. Specifically, cross-dataset experiments are conducted from Waymo to nuScenes while single-dataset adaptation uses KITTI benchmark. Table \ref{tab:datasets} shows an overview of three datasets. Following \cite{yang2021st3d,wang2020train}, we adopt the KITTI evaluation metric for all datasets on the commonly used car category (the vehicle category in Waymo). We report the average precision (AP) over
	40 recall positions. The IoU thresholds are 0.7 for both
	the BEV IoUs and 3D IoUs evaluation. The models are evaluated on the validation set of each dataset.
	\begin{algorithm}[tb]
		\setlength{\belowcaptionskip}{-20pt}
		\caption{Progressive Knowledge Transfer}
		\label{alg:progressive}  
		\begin{algorithmic}[1] 
			\Require Source domain labeled data $\{(X^s_i, Y^s_i)\}^{N_s}_{i=1}$. The (equivalent) beam number and mean point number in each beam of source and target domains are $B_s,P_s$ and $B_t',P_t$ respectively. 
			\Ensure The 3D object detection model for target domain.
			\State Calculate the iteration $n$ for progressive knowledge transfer: $n = \lfloor \log_2 (B_s/B_t') \rfloor$
			\State Pretrain 3D detector $D_0$ on $\{(X^s_i, Y^s_i)\}^{N_s}_{i=1}$.
			\For{$j=1$ to $n$:}
			\State Use the method introduced in Section \ref{sec:pseudo} to downsample $\{X^s_i\}^{N_s}_{i=1}$ to $\{X^{m_j}_i\}^{N_s}_{i=1}$ which have $[ B_s/{2^j}]$ beams. If $j == n$, also downsample the mean point number per beam $P_s$ to $P_t$. 
			\State Adopt $D_{j - 1}$ as the teacher model to train the student model $D_{j}$ on pseudo low-beam data $\{(X^{m_j}_i, Y^s_i)\}^{N_s}_{i=1}$ with the distillation method described in Section \ref{sec:distillation}. 
			\EndFor
			\State $D_n$ is the final model for the target domain. 
			
		\end{algorithmic} 
		
	\end{algorithm}

	\begin{table*}[htbp]
		\centering
		
		\resizebox{0.97\textwidth}{!}{
			\begin{tabular}{c|c|c|c|c}
				\hline
				\multirow{2}{*}{Task} &
				\multirow{2}{*}{Method}  & SECOND-IoU & PV-RCNN & PointPillars \\
				\cline{3-5}
				& & $\text{AP}_{\text{BEV}}$ / $\text{AP}_{\text{3D}}$ &  $\text{AP}_{\text{BEV}}$ / $\text{AP}_{\text{3D}}$ &  $\text{AP}_{\text{BEV}}$ / $\text{AP}_{\text{3D}}$\\
				\hline
				\multirow{8}{*}{Waymo $\rightarrow$ nuScenes} & Direct Transfer  & 32.91 / 17.24 &  34.50 / 21.47 &  27.8 / 12.1 \\
				&SN \cite{wang2020train} & 33.23 / 18.57  &  34.22 / 22.29 & 28.31  / 12.98 \\
				&ST3D & 35.92 / 20.19 &  36.42 / 22.99 & 30.6 / 15.6 \\
				&Hegde \etal \cite{hegde2021attentive} &  \qquad - / 20.47 & -  /  - & - / - \\
				&UMT \cite{hegde2021uncertainty}& 35.10 / 21.05 & - / - & - / - \\
				&3D-CoCo \cite{yihan2021learning}& - / - & - / - & 33.1 / 20.7 \\
				\cline{2-5}
				&Ours &  40.66 / 22.86  &  43.31 / 25.63 & 40.23 / 19.12 \\
				&Ours (w/ ST3D)&  \bf{42.04} / \bf{24.50} &  \bf{44.08} / \bf{26.37} & \bf{40.83}/ \bf{20.97} \\
				\hline
		\end{tabular}}
		
		\caption{Results of cross-dataset adaptation experiments. \textbf{Direct Transfer} indicates that the model trained on the Waymo dataset is directly tested on nuScenes. We report $\text{AP}_{\text{BEV}}$ and $\text{AP}_{\text{3D}}$ over 40 recall positions of the car category at IoU = 0.7. }
		
		\label{tab:cross}
	\end{table*}
	\noindent \textbf{Implementation Details:}
	We evaluate our LiDAR Distillation on three commonly used 3D detectors: PointPillars \cite{lang2019pointpillars}, SECOND \cite{yan2018second} and PV-RCNN \cite{shi2020pv}. Following \cite{yang2021st3d}, we also add an extra IoU head to SECOND, which is named as SECOND-IoU. To guarantee the same input format in each dataset, we only use $(x,y,z)$ as the raw points' features.  To fairly compare with ST3D \cite{yang2021st3d}, we adopt same data augmentation without using GT sampling. The data augmentation is performed simultaneously in a low-beam and high-beam point clouds pair. We utilize 128 ROIs to sample the features on the BEV feature maps and the ratio of positive and negative samples of ROIs was 1:1. The mimic loss weight is set to 1 to balance multiple objective functions. We run full training epochs for each iteration in the progressive knowledge transfer pipeline, \ie, 80 epochs for KITTI and 30 epochs for Waymo. Our work is built upon the 3D object detection codebase OpenPCDet \cite{openpcdet2020}. 

	\subsection{Cross-dataset Adaptation}
	Table \ref{tab:cross} shows the results of cross-dataset experiments. We compare our method with two state-of-the-art general UDA methods SN \cite{wang2020train} and ST3D \cite{yang2021st3d}. The reported numbers of these two methods are from \cite{yang2021st3d}. Although our LiDAR Distillation only aims to bridge the domain gap caused by LiDAR beams and there exist many other domain-variant factors (\eg car sizes, geography and weather) in two datasets, our method still outperforms other general UDA methods for a large margin. The improvements over Direct Transfer baseline are also significant. Moreover, our LiDAR Distillation does not use the training set of the target domain and is complementary to other general UDA methods. Take the ST3D as an example: we can simply replace the model pretrained on the source domain with the model pretrained on pseudo low-beam data.  From Table \ref{tab:cross}, we can see that our method greatly boosts the performance of ST3D ($\text{AP}_{\text{BEV}}$ : 35.92 $\rightarrow$ 42.04), which demonstrates that point cloud density is the key factor among many domain-variant factors.
	
	\begin{table}[tb]
		\centering
		\resizebox{0.7\textwidth}{!}{
			\begin{tabular}{c|c|c|c}
				\hline
				Dataset  & LiDAR Type & VFOV & Points Per Beam\\
				\hline
				Waymo \cite{sun2020scalability} & 64-beam & [-17.6$^{\circ}$, 2.4$^{\circ}$] & 2258 \\
				\hline
				KITTI \cite{Geiger2012KITTI} & 64-beam & [-23.6$^{\circ}$, 3.2$^{\circ}$] & 1863 \\
				\hline
				nuScenes \cite{caesar2020nuscenes} & 32-beam & [-30.0$^{\circ}$, 10.0$^{\circ}$] & 1084 \\
				\hline
		\end{tabular}}
		\caption{Datasets overview. We use \textbf{version 1.0} of Waymo Open Dataset. \textbf{VFOV} means vertical field of view (the range of beam angles). Statistical information is computed from whole dataset. }
		\label{tab:datasets}
	\end{table}

	\subsection{Single-dataset Adaptation}
	To decouple the influence of other factors, we further conduct single-dataset adaptation experiments, where the domain gap is only induced by LiDAR beams. Different
	with research-guided datasets, the training data in industry tends to have similar environments with that during inference, where the environmental domain gaps such as object sizes are marginal. The main gap is the beam difference of LiDARs used during training and inference. Thus, it is valuable to investigate the performance drop only caused by this certain domain gap. Unfortunately,
	the popular autonomous driving datasets \cite{Geiger2012KITTI,caesar2020nuscenes,sun2020scalability} are captured by only one type of LiDAR sensor. To solve the problem, we build the synthetic low-beam target domain by uniformly downsampling the validation set of KITTI. We adopt widely used 3D detector PointPillars \cite{lang2019pointpillars} in this experiment.
	
	Table \ref{tab:single} shows the results of single-dataset adaptation on KITTI dataset. 32$^*$ and 16$^*$  mean that we not only reduce LiDAR beams but also subsample 1/2 points in each beam. We do not compare with SN since car sizes are the same in training and validation sets. We can see that our LiDAR Distillation significantly improves Direct Transfer baseline. As the difference between point cloud densities increases, the performance gap becomes larger. Surprisingly, we find that the general UDA method ST3D \cite{yang2021st3d} will degrade the performance. This is mainly due to different point cloud densities in source and target domains, which are not friendly to the self-training framework. 
	
	\renewcommand{\multirowsetup}{\centering}  
	\begin{table}[tb]
		\centering
		
		\resizebox{1.0\textwidth}{!}{
			\begin{tabular}{c|c|ccc|ccc}
				\hline
				\multirow{3}{2.5cm}{Target Domain \\ Beams} & \multirow{3}{*}{Method}  & \multicolumn{6}{c}{PointPillars}  \\
				\cline{3-8} 
				&  & \multicolumn{3}{c|}{$\text{AP}_{\text{BEV}}$}  & \multicolumn{3}{c}{$\text{AP}_{\text{3D}}$} \\
				
				&  & \multicolumn{1}{c}{Easy} & \multicolumn{1}{c}{Moderate} & \multicolumn{1}{c|}{Hard} & \multicolumn{1}{c}{Easy} & \multicolumn{1}{c}{Moderate} & \multicolumn{1}{c}{Hard} \\
				\hline
				\multirow{4}{*}{$32$} & Direct Transfer  & 91.19 & 79.81 & 76.53 &  80.79 & 65.91 &61.09 \\
				& ST3D \cite{yang2021st3d} & 86.65 &71.29 &66.77 & 75.36 &57.57 &52.88 \\
				\cline{2-8}
				& Ours &\bf{91.59} &\bf{82.22}&\bf{79.57} &  \bf{86.00} & \bf{70.15} &\bf{66.86} \\ 
				&\emph{Improvement} & +0.40 & +2.41& +3.04& +5.21& +4.24& +5.77 \\ 
				\hline
				\hline
				\multirow{4}{*}{$32^{*}$} & Direct Transfer  & 88.39&73.56&68.22&74.59&57.77&51.45 \\
				& ST3D \cite{yang2021st3d} & 83.81 &67.08 &62.57&	71.09&53.30&48.34 \\
				\cline{2-8}
				& Ours &\bf{90.60}&\bf{79.47}&\bf{76.58}&\bf{82.83}&\bf{66.96}&\bf{62.51} \\ 
				&\emph{Improvement} & +2.21 & +5.91& +8.36& +8.24& +9.19& +11.06 \\
				\hline
				\hline
				\multirow{4}{*}{$16$} & Direct Transfer  & 83.11&64.91&58.32&	67.64&47.48&41.41 \\
				& ST3D \cite{yang2021st3d} & 75.49&57.58&52.33&	60.36&42.40&37.93 \\
				\cline{2-8}
				& Ours &\bf{89.98}&\bf{74.32}&\bf{71.54}&\bf{80.21}&\bf{59.87}&\bf{55.32} \\ 
				&\emph{Improvement} & +6.87 & +9.41& +13.22& +12.57& +12.39& +13.91 \\
				\hline
				\hline
				\multirow{4}{*}{$16^{*}$} & Direct Transfer  & 77.22&56.32&49.51&57.36&38.75&32.88 \\
				& ST3D \cite{yang2021st3d} & 76.04&55.63&49.18&55.17&37.02&31.84 \\
				\cline{2-8} 
				& Ours &\bf{87.44}&\bf{70.43}&\bf{66.35}&\bf{75.35}&\bf{55.24}&\bf{50.96} \\ 
				&\emph{Improvement} & +10.22 & +14.11& +16.84& +17.99& +16.49& +18.08 \\
				\hline
		\end{tabular}}
		
		\caption{Results of single-dataset adaptation on KITTI dataset \cite{Geiger2012KITTI}. \textbf{Direct Transfer} indicates that the model trained on the original KITTI training set is directly evaluated on the low-beam validation set. For \textbf{32$^*$} and \textbf{16$^*$}, we not only reduce LiDAR beams but also subsample 1/2 points in each beam. We report $\text{AP}_{\text{BEV}}$ and $\text{AP}_{\text{3D}}$ over 40 recall positions of the car category at IoU = 0.7. The improvement is calculated according to Direct Transfer baseline. We mark the best result of each target domain in \textbf{bold}.}
		\label{tab:single}
	\end{table}
	
	\subsection{Ablation Studies}
	In this section, we conduct sufficient experiments to verify the effectiveness of each proposed component. To exclude the effects of other domain variables, all experiments are conducted under the single-dataset setting with PointPillars \cite{lang2019pointpillars} backbones.
	
	\noindent \textbf{Component Analysis of Teacher-student Framework:} The experiment uses 32-beam data as the target domain. As shown in Table \ref{tab:ablation_main}, all components contribute to the final performance. While high-beam data is informative and we can get high-performance model, it is hard to directly transfer the knowledge to the low-beam domain. With the alignment of the point cloud density, 3D detectors can learn more robust domain-invariant features in the source domain training set. The distillation from real high-beam data will introduce rich knowledge, which can be maximumly preserved in teacher pretrained weights. To better understand the effects of point cloud density alignment and knowledge distillation, we further conduct experiments of all synthetic target domains in supplementary material, where we find that knowledge distillation plays an important role when the domain gap is small while density alignment is more crucial when the domain gap is large. The experiments also show that our
	method significantly outperforms point density alignment
	baseline, i.e. taking the low-beam point clouds as training data, which verify the effectiveness of the proposed teacher-student pipeline.

	\begin{table}[tb]
		\centering
		
		\resizebox{0.7\textwidth}{!}{
			\begin{tabular}{c|c|c|ccc}
				\hline
				density & knowledge  & \multirow{2}{*}{pretrained}& \multicolumn{3}{c}{$\text{AP}_{\text{3D}}$}  \\
				
				alignment& distillation & &\multicolumn{1}{c}{Easy} & \multicolumn{1}{c}{Moderate} & \multicolumn{1}{c}{Hard}   \\
				
				\hline
				& & & 80.79 & 65.91 &61.09 \\
				\hline
				\checkmark & & & 82.80&67.01&63.82 \\
				\hline 
				\checkmark & &\checkmark & 83.05&68.66&64.79\\
				\hline 
				\checkmark & \checkmark& &83.71&69.01&64.97 \\
				\hline 
				\checkmark & \checkmark& \checkmark&\bf{86.00}&\bf{70.15}&\bf{66.86} \\
				\hline
				
		\end{tabular}}
		
		\caption{The ablation study of the teacher-student framework. \textbf{Density alignment} means that we use generated low-beam data to train the network. \textbf{Pretrained} indicates that we employ pretrained teacher weights as the initial weights for the student network. 32-beam data is set as the target domain. }
		\label{tab:ablation_main}
	\end{table}

	\begin{table}[htbp]
		\begin{minipage}{0.49\linewidth}
			\centering
			
			\begin{tabular}{c|ccc}
				\hline
				\multirow{2}{*}{teacher model}& \multicolumn{3}{c}{$\text{AP}_{\text{3D}}$}  \\
				
				&\multicolumn{1}{c}{Easy} & \multicolumn{1}{c}{Moderate} & \multicolumn{1}{c}{Hard}   \\
				
				\hline
				$64$ & 78.75&58.80&54.47 \\
				\hline
				$32$& 78.31&58.74&\bf{55.62}\\
				\hline 
				32 progressive & \bf{80.21} & \bf{59.87} &55.32 \\ 
				\hline
				
			\end{tabular}
			
			\caption{The ablation study of progressive knowledge transfer. \textbf{64} and \textbf{32} represent that the teacher model is trained without progressive distillation. The teacher model obtained from our method is named as \textbf{32 progressive}. The result is evaluated on 16-beam data.}
			\label{tab:ablation_progressive}
		\end{minipage}
		\begin{minipage}{0.49\linewidth}
			\centering
			
			\begin{tabular}{c|ccc}
				\hline
				\multirow{2}{*}{mimicking regions}& \multicolumn{3}{c}{$\text{AP}_{\text{3D}}$}  \\
				
				&\multicolumn{1}{c}{Easy} & \multicolumn{1}{c}{Moderate} & \multicolumn{1}{c}{Hard}   \\
				
				\hline
				all & 83.83&69.36&64.98 \\
				\hline
				groundtruth& 84.46&69.99&65.41\\
				\hline 
				ROI & \bf{86.00}&\bf{70.15}&\bf{66.86}  \\
				\hline
				
			\end{tabular}
			
			\caption{The ablation study of mimicking regions. \textbf{all} means that we perform mimicking operation on the whole BEV feature maps while \textbf{groundtruth} indicates that we only mimic the regions inside groundtruth bounding boxes.}
			
			\label{tab:ablation_mimic}
		\end{minipage}
	\end{table}
	
	\noindent \textbf{Effectiveness of Progressive Knowledge Transfer:} We further investigate the effectiveness of progressive knowledge transfer. From Table \ref{tab:ablation_progressive}, we can see that if the teacher model is trained on original 64-beam data, it cannot provide effective guidance for the student model since there exist large gaps between 64-beam and 16-beam data. Moreover, compared with the teacher model directly trained on 32-beam data, the teacher network obtained from the proposed progressive pipeline gets better results.

	\noindent \textbf{Analysis for Mimicking Regions:}
	The experimental result is shown in Table \ref{tab:ablation_mimic}. We find that it is not effective to mimic all elements of BEV feature maps. We also try mimicking the regions inside groundtruth bounding boxes. However, the results are not promising since some background features are also useful. With a good balance of foreground and background information, mimicking ROI regions performs best.  
	
	\subsection{Further Analysis}
	\noindent \textbf{The Necessity of Point Cloud Density Alignment:}
	In this part, we study whether the point cloud density is a key factor to bridge the domain gap among many domain-variant factors. Although the data in nuScenes \cite{caesar2020nuscenes} is 32-beam, according to the VFOV in Table \ref{tab:datasets} and Equation \ref{eq:beam}, its equivalent beam in the source domain is 16. In addition, the points per beam of Waymo data is about twice as much as that of nuScenes data. Therefore, 16$^*$ beam is the most similar point cloud density compared with target domain data. Experimental results are shown in Table \ref{tab:key}, where evaluation results on target and source domain are illustrated in the first and second columns respectively. We observe that we can get better results when point cloud densities of source and target domain become closer. Indeed, both the beams of Waymo and nuScenes obtain non-uniform distribution and our method does not require them to strictly align. While the alignment of the point cloud density will sacrifice the performance on the source domain due to the information loss, it will bring huge improvements on the target domain. 
	
	\noindent \textbf{Comparison with LiDAR Upsampling Methods:}
	An alternative solution for point cloud density alignment is to upsample the low-beam data in the target domain. Different from scene completion methods \cite{yan2021sparse,roldao2020lmscnet} which utilize 3D convolutions, we do not need to complete the whole scenes but need to process point clouds efficiently. To solve the problem, we first convert point clouds to range images. Then we upsample range images by naive non-parameter bilinear interpolation and a super-resolution network \cite{Hui-IMDN-2019}. As shown in Table \ref{tab:upsample}, these two upsampling methods fail to boost the accuracy, which demonstrates that it is not trivial to accurately upsample LiDAR points with high efficiency. This is mainly because there exist some noisy points, which will lead to incorrect object shapes. 
	\begin{table}[tb]
		\begin{minipage}{0.48\linewidth}
			\centering
			
			\resizebox{1.0\textwidth}{!}{
				\begin{tabular}{c|c|c}
					\hline
					Source Domain& \multicolumn{2}{c}{$\text{AP}_{\text{BEV}}$/$\text{AP}_{\text{3D}}$}  \\
					\cline{2-3}
					
					Beams& nuScenes & Waymo  \\
					
					\hline
					$64$ & 32.91/17.24& \bf{67.72}/\bf{54.01} \\
					\hline
					$32$ & 37.35/20.66& 64.67/51.18 \\
					\hline 
					$16$ & 40.08/21.51& 57.04/43.41 \\
					\hline 
					$16^*$ & \bf{40.66}/\bf{22.86} & 53.23/39.97 \\
					\hline 
			\end{tabular}}
			
			\caption{The Necessity of Point Cloud Density Alignment. We downsample Waymo \cite{sun2020scalability} data to different beams and evaluate models on nuScenes \cite{caesar2020nuscenes} (target domain) and Waymo \cite{sun2020scalability} (source domain). Notice that the equivalent beam of nuScenes is 16$^*$. The experiment is conducted with SECOND-IoU backbone.} 
			
			\label{tab:key}
		\end{minipage}
		\begin{minipage}{0.48\linewidth}
			\centering
			
			\resizebox{1.0\textwidth}{!}{
				\begin{tabular}{c|ccc}
					\hline
					\multirow{2}{*}{Method}& \multicolumn{3}{c}{$\text{AP}_{\text{3D}}$}  \\
					
					&\multicolumn{1}{c}{Easy} & \multicolumn{1}{c}{Moderate} & \multicolumn{1}{c}{Hard}   \\
					\hline
					interpolation & 74.43&54.56&49.72\\
					\hline
					super resolution& 79.78&62.58&58.04\\
					\hline 
					Ours & \bf{86.00}&\bf{70.15}&\bf{66.86} \\ 
					\hline
					
			\end{tabular}}
			
			\caption{Comparison with LiDAR upsampling methods. We first convert point clouds of the target domain to range images. Then we upsample range images with bilinear interpolation and a light-weight super-resolution network \cite{Hui-IMDN-2019}, and convert them back to point clouds. 32-beam KITTI data is used as the target domain.}
			\label{tab:upsample}
		\end{minipage}
	\end{table}

	\section{Conclusion}
	In this work, we propose the LiDAR Distillation to bridge the domain gap caused by different LiDAR beams in 3D object detection task. Inspired by the discovery that point cloud density is an important factor for this problem, we first generate pseudo low-beam data by downsampling real high-beam LiDAR points. Then knowledge distillation technique is used to progressively boost the performance of the model trained on synthetic low-beam point clouds. Experimental results on three popular autonomous driving datasets demonstrate the effectiveness of our method. 
	
	\section*{Acknowledgments}
	This work was supported in part by the National Key Research and Development Program of China under Grant 2017YFA0700802, in part by the National Natural Science Foundation of China under Grant 62125603 and Grant U1813218, in part by a grant from the Beijing Academy of Artificial Intelligence (BAAI).
	
	\clearpage
	%
	%
	\bibliographystyle{splncs04}
	\bibliography{egbib}

\begin{thebibliography}{10}
\providecommand{\url}[1]{\texttt{#1}}
\providecommand{\urlprefix}{URL }
\providecommand{\doi}[1]{https://doi.org/#1}

\bibitem{anil2018large}
Anil, R., Pereyra, G., Passos, A., Ormandi, R., Dahl, G.E., Hinton, G.E.: Large
  scale distributed neural network training through online distillation. arXiv
  preprint arXiv:1804.03235  (2018)

\bibitem{caesar2020nuscenes}
Caesar, H., Bankiti, V., Lang, A.H., Vora, S., Liong, V.E., Xu, Q., Krishnan,
  A., Pan, Y., Baldan, G., Beijbom, O.: nuscenes: A multimodal dataset for
  autonomous driving. In: CVPR. pp. 11621--11631 (2020)

\bibitem{chen2020harmonizing}
Chen, C., Zheng, Z., Ding, X., Huang, Y., Dou, Q.: Harmonizing transferability
  and discriminability for adapting object detectors. In: CVPR. pp. 8869--8878
  (2020)

\bibitem{chen2017learning}
Chen, G., Choi, W., Yu, X., Han, T., Chandraker, M.: Learning efficient object
  detection models with knowledge distillation. NeurIPS  (2017)

\bibitem{chen2017multi}
Chen, X., Ma, H., Wan, J., Li, B., Xia, T.: Multi-view 3d object detection
  network for autonomous driving. In: CVPR. pp. 1907--1915 (2017)

\bibitem{chen2018domain}
Chen, Y., Li, W., Sakaridis, C., Dai, D., Van~Gool, L.: Domain adaptive faster
  r-cnn for object detection in the wild. In: CVPR. pp. 3339--3348 (2018)

\bibitem{choy20194d}
Choy, C., Gwak, J., Savarese, S.: 4d spatio-temporal convnets: Minkowski
  convolutional neural networks. In: CVPR. pp. 3075--3084 (2019)

\bibitem{deng2021voxel}
Deng, J., Shi, S., Li, P., Zhou, W., Zhang, Y., Li, H.: Voxel r-cnn: Towards
  high performance voxel-based 3d object detection. In: AAAI. pp. 1201--1209
  (2021)

\bibitem{Geiger2012KITTI}
Geiger, A., Lenz, P., Urtasun, R.: Are we ready for autonomous driving? the
  kitti vision benchmark suite. In: CVPR (2012)

\bibitem{graham20183d}
Graham, B., Engelcke, M., Van Der~Maaten, L.: 3d semantic segmentation with
  submanifold sparse convolutional networks. In: CVPR. pp. 9224--9232 (2018)

\bibitem{guo2021distilling}
Guo, J., Han, K., Wang, Y., Wu, H., Chen, X., Xu, C., Xu, C.: Distilling object
  detectors via decoupled features. In: CVPR. pp. 2154--2164 (2021)

\bibitem{hegde2021attentive}
Hegde, D., Patel, V.: Attentive prototypes for source-free unsupervised domain
  adaptive 3d object detection. arXiv preprint arXiv:2111.15656  (2021)

\bibitem{hegde2021uncertainty}
Hegde, D., Sindagi, V., Kilic, V., Cooper, A.B., Foster, M., Patel, V.:
  Uncertainty-aware mean teacher for source-free unsupervised domain adaptive
  3d object detection. arXiv preprint arXiv:2109.14651  (2021)

\bibitem{heo2019comprehensive}
Heo, B., Kim, J., Yun, S., Park, H., Kwak, N., Choi, J.Y.: A comprehensive
  overhaul of feature distillation. In: ICCV. pp. 1921--1930 (2019)

\bibitem{hinton2015distilling}
Hinton, G., Vinyals, O., Dean, J.: Distilling the knowledge in a neural
  network. arXiv preprint arXiv:1503.02531  (2015)

\bibitem{Hui-IMDN-2019}
Hui, Z., Gao, X., Yang, Y., Wang, X.: Lightweight image super-resolution with
  information multi-distillation network. In: ACMMM. pp. 2024--2032 (2019)

\bibitem{jaritz2020xmuda}
Jaritz, M., Vu, T.H., Charette, R.d., Wirbel, E., P{\'e}rez, P.: xmuda:
  Cross-modal unsupervised domain adaptation for 3d semantic segmentation. In:
  CVPR. pp. 12605--12614 (2020)

\bibitem{lyft2019}
Kesten, R., Usman, M., Houston, J., Pandya, T., Nadhamuni, K., Ferreira, A.,
  Yuan, M., Low, B., Jain, A., Ondruska, P., Omari, S., Shah, S., Kulkarni, A.,
  Kazakova, A., Tao, C., Platinsky, L., Jiang, W., Shet, V.: Lyft level 5
  perception dataset 2020. \url{https://level5.lyft.com/dataset/} (2019)

\bibitem{ku2018joint}
Ku, J., Mozifian, M., Lee, J., Harakeh, A., Waslander, S.L.: Joint 3d proposal
  generation and object detection from view aggregation. In: IROS. pp.~1--8
  (2018)

\bibitem{lang2019pointpillars}
Lang, A.H., Vora, S., Caesar, H., Zhou, L., Yang, J., Beijbom, O.:
  Pointpillars: Fast encoders for object detection from point clouds. In: CVPR.
  pp. 12697--12705 (2019)

\bibitem{li2020spatialattention}
Li, C., Du, D., Zhang, L., Wen, L., Luo, T., Wu, Y., Zhu, P.: Spatial attention
  pyramid network for unsupervised domain adaptation. In: ECCV. pp. 481--497
  (2020)

\bibitem{li2017mimicking}
Li, Q., Jin, S., Yan, J.: Mimicking very efficient network for object
  detection. In: CVPR. pp. 6356--6364 (2017)

\bibitem{luo2021unsupervised}
Luo, Z., Cai, Z., Zhou, C., Zhang, G., Zhao, H., Yi, S., Lu, S., Li, H., Zhang,
  S., Liu, Z.: Unsupervised domain adaptive 3d detection with multi-level
  consistency. In: ICCV. pp. 8866--8875 (2021)

\bibitem{milioto2019rangenet++}
Milioto, A., Vizzo, I., Behley, J., Stachniss, C.: Rangenet++: Fast and
  accurate lidar semantic segmentation. In: 2019 IEEE/RSJ International
  Conference on Intelligent Robots and Systems (IROS). pp. 4213--4220. IEEE
  (2019)

\bibitem{qi2018frustum}
Qi, C.R., Liu, W., Wu, C., Su, H., Guibas, L.J.: Frustum pointnets for 3d
  object detection from rgb-d data. In: CVPR. pp. 918--927 (2018)

\bibitem{qin2019pointdan}
Qin, C., You, H., Wang, L., Kuo, C.C.J., Fu, Y.: Pointdan: A multi-scale 3d
  domain adaption network for point cloud representation. In: NeurIPS. pp.
  7192--7203 (2019)

\bibitem{roldao2020lmscnet}
Rold{\~a}o, L., de~Charette, R., Verroust-Blondet, A.: Lmscnet: Lightweight
  multiscale 3d semantic completion. In: 3DV (2020)

\bibitem{romero2014fitnets}
Romero, A., Ballas, N., Kahou, S.E., Chassang, A., Gatta, C., Bengio, Y.:
  Fitnets: Hints for thin deep nets. arXiv preprint arXiv:1412.6550  (2014)

\bibitem{saito2017asymmetric}
Saito, K., Ushiku, Y., Harada, T.: Asymmetric tri-training for unsupervised
  domain adaptation. In: ICML. pp. 2988--2997 (2017)

\bibitem{saito2019strongweak}
Saito, K., Ushiku, Y., Harada, T., Saenko, K.: Strong-weak distribution
  alignment for adaptive object detection. In: CVPR. pp. 6956--6965 (2019)

\bibitem{saito2018maximum}
Saito, K., Watanabe, K., Ushiku, Y., Harada, T.: Maximum classifier discrepancy
  for unsupervised domain adaptation. In: CVPR. pp. 3723--3732 (2018)

\bibitem{saltori2020sf}
Saltori, C., Lathuili{\`e}re, S., Sebe, N., Ricci, E., Galasso, F.: {SF-UDA 3D:
  Source-Free Unsupervised Domain Adaptation for LiDAR-Based 3D Object
  Detection}. In: 3DV. pp. 771--780 (2020)

\bibitem{shi2020pv}
Shi, S., Guo, C., Jiang, L., Wang, Z., Shi, J., Wang, X., Li, H.: Pv-rcnn:
  Point-voxel feature set abstraction for 3d object detection. In: CVPR. pp.
  10529--10538 (2020)

\bibitem{shi2019pointrcnn}
Shi, S., Wang, X., Li, H.: Pointrcnn: 3d object proposal generation and
  detection from point cloud. In: CVPR. pp. 770--779 (2019)

\bibitem{shi2019points}
Shi, S., Wang, Z., Shi, J., Wang, X., Li, H.: From points to parts: 3d object
  detection from point cloud with part-aware and part-aggregation network.
  arXiv preprint arXiv:1907.03670  (2019)

\bibitem{sun2020scalability}
Sun, P., Kretzschmar, H., Dotiwalla, X., Chouard, A., Patnaik, V., Tsui, P.,
  Guo, J., Zhou, Y., Chai, Y., Caine, B., et~al.: Scalability in perception for
  autonomous driving: Waymo open dataset. In: CVPR. pp. 2446--2454 (2020)

\bibitem{openpcdet2020}
Team, O.D.: Openpcdet: An open-source toolbox for 3d object detection from
  point clouds. \url{https://github.com/open-mmlab/OpenPCDet} (2020)

\bibitem{tung2019similarity}
Tung, F., Mori, G.: Similarity-preserving knowledge distillation. In: ICCV. pp.
  1365--1374 (2019)

\bibitem{wang2019distilling}
Wang, T., Yuan, L., Zhang, X., Feng, J.: Distilling object detectors with
  fine-grained feature imitation. In: CVPR. pp. 4933--4942 (2019)

\bibitem{wang2020train}
Wang, Y., Chen, X., You, Y., Li, L.E., Hariharan, B., Campbell, M., Weinberger,
  K.Q., Chao, W.L.: Train in germany, test in the usa: Making 3d object
  detectors generalize. In: CVPR. pp. 11713--11723 (2020)

\bibitem{wei2018quantization}
Wei, Y., Pan, X., Qin, H., Ouyang, W., Yan, J.: Quantization mimic: Towards
  very tiny cnn for object detection. In: ECCV. pp. 267--283 (2018)

\bibitem{wu2018squeezeseg}
Wu, B., Wan, A., Yue, X., Keutzer, K.: Squeezeseg: Convolutional neural nets
  with recurrent crf for real-time road-object segmentation from 3d lidar point
  cloud. In: 2018 IEEE International Conference on Robotics and Automation
  (ICRA). pp. 1887--1893. IEEE (2018)

\bibitem{wu2019squeezesegv2}
Wu, B., Zhou, X., Zhao, S., Yue, X., Keutzer, K.: Squeezesegv2: Improved model
  structure and unsupervised domain adaptation for road-object segmentation
  from a lidar point cloud. In: ICRA. pp. 4376--4382 (2019)

\bibitem{xu2020exploring}
Xu, C.D., Zhao, X.R., Jin, X., Wei, X.S.: Exploring categorical regularization
  for domain adaptive object detection. In: CVPR. pp. 11724--11733 (2020)

\bibitem{xu2021spg}
Xu, Q., Zhou, Y., Wang, W., Qi, C.R., Anguelov, D.: {SPG: Unsupervised Domain
  Adaptation for 3D Object Detection via Semantic Point Generation}. In: ICCV.
  pp. 15446--15456 (2021)

\bibitem{yan2021sparse}
Yan, X., Gao, J., Li, J., Zhang, R., Li, Z., Huang, R., Cui, S.: Sparse single
  sweep lidar point cloud segmentation via learning contextual shape priors
  from scene completion. In: AAAI. pp. 3101--3109 (2021)

\bibitem{yan2018second}
Yan, Y., Mao, Y., Li, B.: Second: Sparsely embedded convolutional detection.
  Sensors  \textbf{18}(10), ~3337 (2018)

\bibitem{yang2018pixor}
Yang, B., Luo, W., Urtasun, R.: Pixor: Real-time 3d object detection from point
  clouds. In: CVPR. pp. 7652--7660 (2018)

\bibitem{yang2021st3d}
Yang, J., Shi, S., Wang, Z., Li, H., Qi, X.: {ST3D: Self-training for
  Unsupervised Domain Adaptation on 3D Object Detection}. In: CVPR. pp.
  10368--10378 (2021)

\bibitem{yang2019std}
Yang, Z., Sun, Y., Liu, S., Shen, X., Jia, J.: Std: Sparse-to-dense 3d object
  detector for point cloud. In: ICCV. pp. 1951--1960 (2019)

\bibitem{yi2021complete}
Yi, L., Gong, B., Funkhouser, T.: Complete \& label: A domain adaptation
  approach to semantic segmentation of lidar point clouds. In: CVPR. pp.
  15363--15373 (2021)

\bibitem{yihan2021learning}
Yihan, Z., Wang, C., Wang, Y., Xu, H., Ye, C., Yang, Z., Ma, C.: Learning
  transferable features for point cloud detection via 3d contrastive
  co-training. NeurIPS  \textbf{34} (2021)

\bibitem{yim2017gift}
Yim, J., Joo, D., Bae, J., Kim, J.: A gift from knowledge distillation: Fast
  optimization, network minimization and transfer learning. In: CVPR. pp.
  4133--4141 (2017)

\bibitem{you2017learning}
You, S., Xu, C., Xu, C., Tao, D.: Learning from multiple teacher networks. In:
  SIGKDD. pp. 1285--1294 (2017)

\bibitem{zhang2021srdan}
Zhang, W., Li, W., Xu, D.: Srdan: Scale-aware and range-aware domain adaptation
  network for cross-dataset 3d object detection. In: CVPR. pp. 6769--6779
  (2021)

\bibitem{zheng2020crossdomain}
Zheng, Y., Huang, D., Liu, S., Wang, Y.: Cross-domain object detection through
  coarse-to-fine feature adaptation. In: CVPR. pp. 13766--13775 (2020)

\bibitem{zhou2018unsupervised}
Zhou, X., Karpur, A., Gan, C., Luo, L., Huang, Q.: Unsupervised domain
  adaptation for 3d keypoint estimation via view consistency. In: ECCV. pp.
  137--153 (2018)

\bibitem{zhou2018voxelnet}
Zhou, Y., Tuzel, O.: Voxelnet: End-to-end learning for point cloud based 3d
  object detection. In: Proceedings of the IEEE Conference on Computer Vision
  and Pattern Recognition. pp. 4490--4499 (2018)

\bibitem{zhu2019selective}
Zhu, X., Pang, J., Yang, C., Shi, J., Lin, D.: Adapting object detectors via
  selective cross-domain alignment. In: CVPR. pp. 687--696 (2019)

\bibitem{zou2018unsupervised}
Zou, Y., Yu, Z., Vijaya~Kumar, B., Wang, J.: Unsupervised domain adaptation for
  semantic segmentation via class-balanced self-training. In: ECCV. pp.
  289--305 (2018)

\end{thebibliography}
	
	\clearpage
	\section*{Appendix}
	\appendix

	\renewcommand{\multirowsetup}{\centering}  
	\begin{table*}[htbp]
		\centering
		\resizebox{0.95\textwidth}{!}{
			\begin{tabular}{c|c|c|ccc|ccc}
				\hline
				\multirow{3}{2cm}{Target Domain \\ Beams} & \multirow{3}{1.5cm}{density \\ alignment}  &
				\multirow{3}{1.5cm}{knowledge \\ distillation}  &
				\multicolumn{6}{c}{PointPillars}  \\
				\cline{4-9} 
				& & & \multicolumn{3}{c|}{$\text{AP}_{\text{3D}}$}  & \multicolumn{3}{c}{\emph{\small{Improvement}}} \\
				
				&  & &\multicolumn{1}{c}{Easy} & \multicolumn{1}{c}{Moderate} & \multicolumn{1}{c|}{Hard} & \multicolumn{1}{c}{Easy} & \multicolumn{1}{c}{Moderate} & \multicolumn{1}{c}{Hard} \\
				\hline
				\multirow{3}{*}{$32$} & &  & 80.79 & 65.91 &61.09 &  - & - &- \\
				& \checkmark & & 82.80 &67.01 &63.82 & +2.01 & +1.10 & +2.73\\
				& \checkmark & \checkmark & 86.00& 70.15& 66.86 & +3.20 & +3.14 & +3.04 \\
				\hline
				\hline
				\multirow{3}{*}{$32^*$} & &  & 74.59&57.77&51.45 &  - & - &- \\
				& \checkmark & & 78.74&63.02&58.94 & +4.15 & +5.25 & +7.49\\
				& \checkmark & \checkmark & 82.83&66.96&62.51 & +4.09 & +3.94 & +3.57 \\
				\hline
				\hline
				\multirow{3}{*}{$16$} & &   & 67.64&47.48&41.41 &  - & - &- \\
				& \checkmark & & 76.12&57.75&53.85 & +8.48 & +10.27 & +12.44 \\
				& \checkmark & \checkmark & 80.21&59.87&55.32 & +4.09 & +2.12 & +1.47 \\
				\hline
				\hline
				\multirow{3}{*}{$16^*$} & &  & 57.36&38.75&32.88 &  - & - &- \\
				& \checkmark & & 70.70&51.24&47.60 & +13.34 & +12.49 & +14.72 \\
				& \checkmark & \checkmark & 75.35&55.24&50.96 & +4.65 & +4.00 & +3.36 \\
				\hline
		\end{tabular}}
		\caption{Component analysis on all target domains of KITTI dataset \cite{Geiger2012KITTI}. For \textbf{32$^*$} and \textbf{16$^*$}, we not only reduce LiDAR beams but also subsample 1/2 points in each beam. The improvement is calculated in a progressive way.}
		\label{tab:ablation}
		\vspace{-10mm}
	\end{table*}
	
	\begin{table}[htbp]
		\centering
		
		\resizebox{0.8\textwidth}{!}{
			\begin{tabular}{c|c|c}
				\hline
				\multirow{2}{*}{Task} &
				\multirow{2}{*}{Method}  & PointPillars \\
				\cline{3-3}
				& & $\text{AP}_{\text{BEV}}$ / $\text{AP}_{\text{3D}}$ \\
				\hline
				\multirow{4}{*}{KITTI $\rightarrow$ nuScenes} & Direct Transfer  & 7.86 / 1.05\\
				&SN\cite{wang2020train} & 14.96 / 5.28 \\
				&ST3D\cite{yang2021st3d} & 19.49 / 6.63 \\
				&Ours (naive downsample) &  20.63 / 7.93 \\
				&Ours &  \bf{21.90} / \bf{9.25} \\
				\hline
		\end{tabular}}
		\caption{Results of KITTI $\rightarrow$ nuScenes adaptation. }
		\vspace{-10mm}
		\label{tab:kitti_nus}
	\end{table}
	
	\section{More Dataset and Implementation Details}
	As a popular 3D object detection benchmark, KITTI \cite{Geiger2012KITTI} contains 3,712 training samples and 3,769 validation samples. Since KITTI dataset only  provide the 3D bounding box labels for the objects within the field of view of the front RGB
	camera, we remove points outside of the front regions both for training and evaluation. According to the
	occlusion, truncation and 2D bounding box height, the objects are divided into three difficulty levels (Easy, Moderate and Hard). 
	
	The Waymo Open Dataset \cite{sun2020scalability} is a large-scale dataset, which contains
	1000 sequences in total, including 798 sequences (158,081 frames) in the training set and 202 sequences (39,987 frames) in the validation
	set. We used 1.0 version of Waymo Open Dataset. Same to ST3D \cite{yang2021st3d}, we also subsampled 1/2 training samples. Note that Waymo data is captured by a 64-beam LiDAR and 4 200-beam short-range LiDAR. The 200-beam LiDAR only captures data in a limited range and most of points come from 64-beam LiDAR. Thus, we only downsampled the points from 64-beam LiDAR.
	
	The nuScenes dataset \cite{caesar2020nuscenes} consists of 28,130 training samples and 6,019 validation samples. The point clouds in nuScenes are 32-beam data  while the equivalent beam to Waymo is 16$^*$. We only used nuScenes for evaluation. 
	
	The voxel size of SECOND and PV-RCNN are set to $(0.05m, 0.05m, 0.1m)$ on KITTI dataset and  $(0.1m, 0.1m, 0.15m)$ on Waymo and nuScenes datasets. The models are trained on 8 RTX 2080 Tis. 
	
	\section{Component Analysis on Different Target Domains}
	To better understand the effects of point cloud density alignment and knowledge distillation, we conduct ablation studies on all synthetic target domains of KITTI \cite{Geiger2012KITTI} dataset. Table \ref{tab:ablation} shows the results. We observe that both point cloud density alignment and knowledge distillation contribute to the final results. On the one hand, when the domain gap is not large (\eg 64 $\rightarrow$ 32), the knowledge distillation plays an more important role. On the other hand, when the domain gap is huge (\eg 64 $\rightarrow$ 16$^*$), it is more crucial to align the point cloud density. 
	
	\section{Additional cross-dataset adaptation}
	As mentioned in ST3D \cite{yang2021st3d}, KITTI dataset lacks of ring view
	annotations and sufficient data. Due to these reasons,
	few methods select KITTI as source domain. To further
	demonstrate the effectiveness of our method, we add KITTI
	$\rightarrow$ nuScenes experiments with PointPillars backbone. During inference, we use the same field of view with that in KITTI dataset. However, we find that environmental domain gaps (such as object sizes) between these two datasets are huge and only using ST3D or our method cannot work well. Thus we combine ST3D and SN with our method and ST3D is also combined with SN. As shown in Table \ref{tab:kitti_nus}, our method can boost the performance of state-of-the-art methods with large margins. We also did ablation study on point cloud dowmsampling methods and we find that the proposed pseudo low-beam data generation method is better than naive downsampling method.

	\section{The value to industry application} We finetune the PointPillars models on nuScenes datasets with different amount of groundtruth, which are pretrained with Waymo $\rightarrow$ nuScenes adaptation. In Table \ref{tab:finetune}, the model pretrained with our method outperforms than other methods. With less groundtruth, the performance gains become larger. Surprisingly, with only 5\% data, we can get higher performance than the model trained from scratch with 100\% data. This experiment shows that we can use our method to reduce the need of expensive 3D labels, which is valuable to the industry. 
	\renewcommand{\multirowsetup}{\centering} 
	\begin{table}[htbp]
		
		\centering
		\resizebox{0.8\textwidth}{!}{
			\begin{tabular}{c|c|c|c}
				\hline
				\multirow{2}{2cm}{Pretrained \\ Method}  &\multicolumn{3}{c}{$\text{AP}_{\text{BEV}}$ / $\text{AP}_{\text{3D}}$} \\
				\cline{2-4}
				& 5\% & 10\% & 100\% \\
				\hline
				Scratch  & 23.77 / 8.07 &  30.60 / 13.78 &  45.31 / 25.84 \\
				Direct Transfer  & 40.60 / 21.50 &  43.34 / 23.77 &  48.74 / 27.06 \\
				ST3D & 43.66 / 24.03 &  45.98 / 25.72 & 49.70 / 29.34 \\
				Ours &  \bf{47.16} / \bf{26.57} &  \bf{49.10} / \bf{28.73} & \bf{51.95}/ \bf{31.34} \\
				\hline
		\end{tabular}}
		\caption{Results of finetuning experiments on nuScenes dataset.}
		\label{tab:finetune}
	\end{table}
	
	\begin{figure}[tb]
		\centering
		\setlength\tabcolsep{1.0pt} 
		\renewcommand{\arraystretch}{1.0}
		\begin{tabular}{ccc}
			{\includegraphics[width=0.33\linewidth]{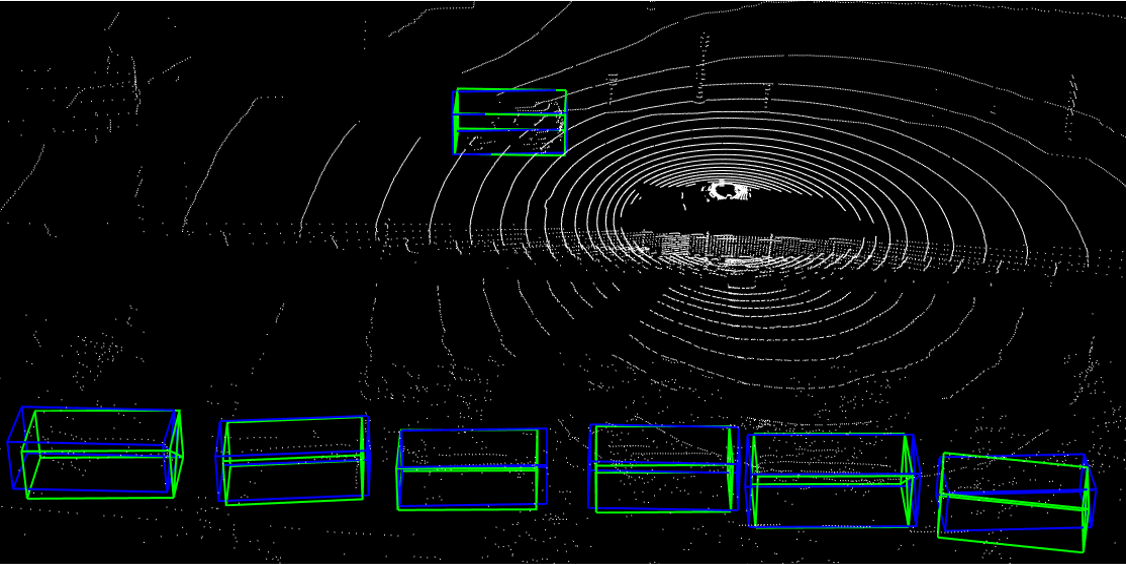}} &
			{\includegraphics[width=0.33\linewidth]{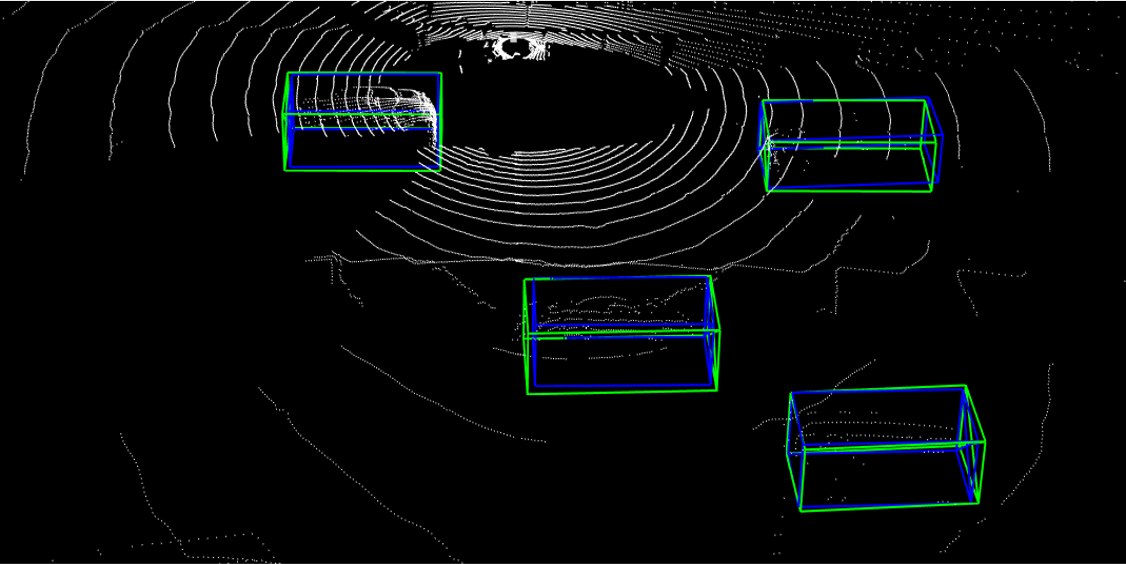}} &
			{\includegraphics[width=0.33\linewidth]{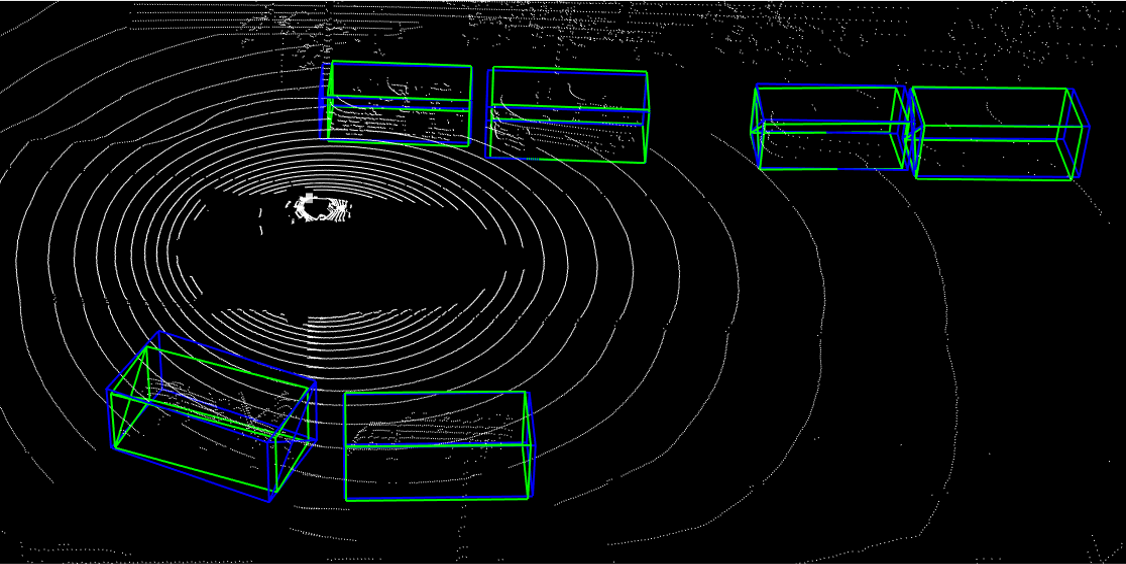}} \\
			{\includegraphics[width=0.33\linewidth]{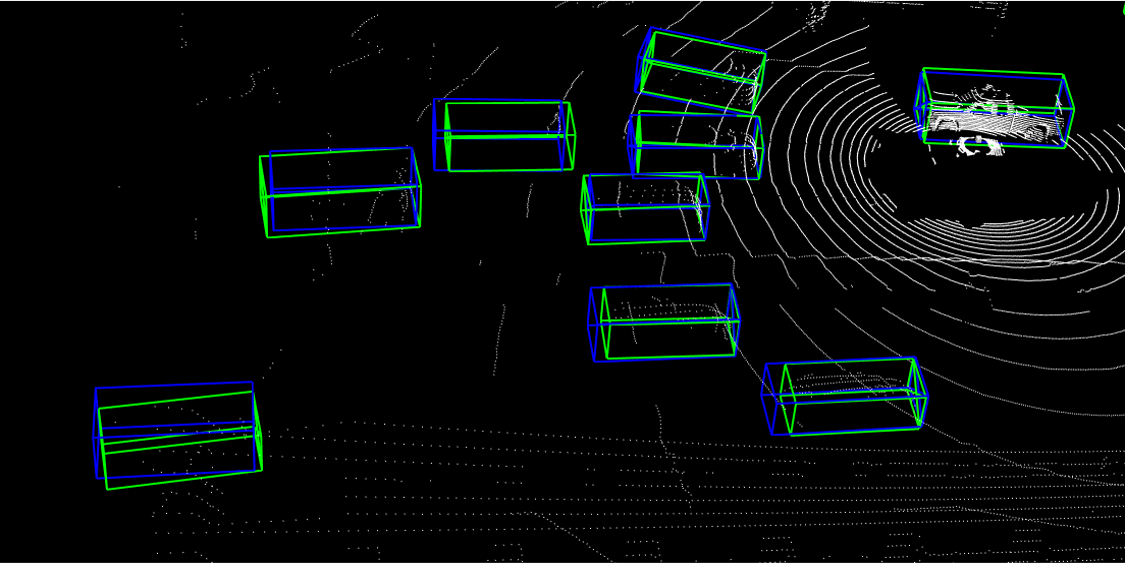}} &
			{\includegraphics[width=0.33\linewidth]{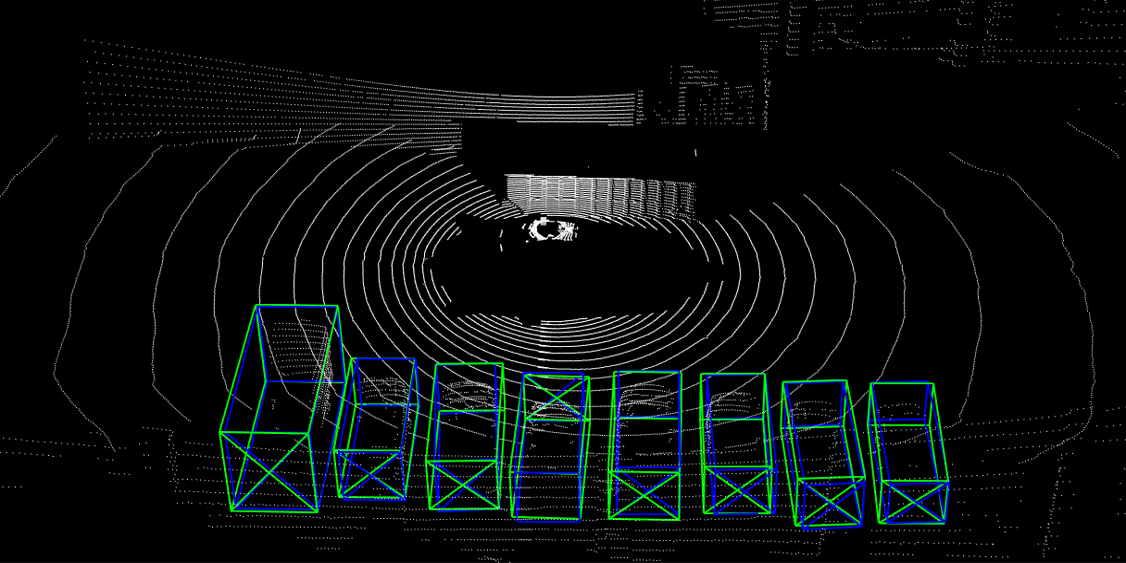}} &
			{\includegraphics[width=0.33\linewidth]{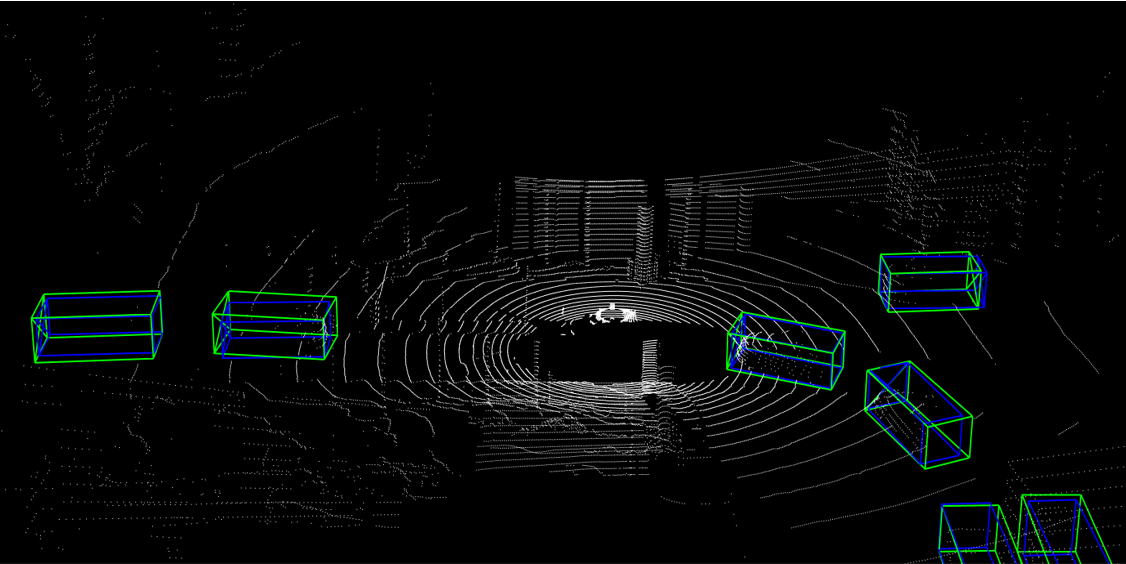}} \\
		\end{tabular}
		\centering
		\caption{Qualitative results of Waymo $\rightarrow$ nuScenes adaptation task. The green and blue bounding boxes
			represent detector predictions and groundtruths respectively.}
		\label{fig}
	\end{figure}
	
	\section{Qualitative Results}
	To better illustrate the superiority of our method, we finally provide some visualizations. Figure \ref{fig} shows qualitative results of cross-dataset adaptation equipped with SECOND-IoU \cite{yan2018second}. We can see that our method can predict high-quality 3D bounding boxes.
	
\end{document}